\documentclass[letterpaper, 10 pt, journal, twoside]{IEEEtran}
\usepackage{makecell}
\usepackage{booktabs}
\usepackage{array}
\newcolumntype{P}[1]{>{\centering\arraybackslash}m{#1}}

\usepackage{graphics}
\usepackage{bbm}
\usepackage[pdftex]{graphicx}
\usepackage{wrapfig}
\DeclareGraphicsExtensions{.pdf,.png,.jpg}
\usepackage{epsfig}
\usepackage[font={small}]{caption}
\usepackage[font=normalsize,subrefformat=parens]{subcaption}
\usepackage[rightcaption]{sidecap}
\usepackage{pbox}

\usepackage{bigstrut}
\usepackage{mathtools}
\usepackage{amsmath, amssymb, amscd}
\usepackage{ wasysym } 
\usepackage{amsfonts}
\usepackage{mathptmx} 
\usepackage{gensymb} 
\usepackage{nicefrac}       
\numberwithin{equation}{section} 
\usepackage{siunitx}
\DeclareMathAlphabet{\mathcal}{OMS}{lmsy}{m}{n}
\DeclareSymbolFont{largesymbols}{OMX}{cmex}{m}{n}
\usepackage{textcomp} 

\usepackage{algorithm} 
\usepackage[noend]{algorithmic} 

\usepackage{array} 
\usepackage{tabularx}
\usepackage{multirow}
\usepackage{booktabs}
\usepackage{tabulary}

\usepackage[T1]{fontenc} 
\usepackage[utf8]{inputenc}
\usepackage[english]{babel} 
\usepackage{units}
\usepackage{bm}
\usepackage{times} 
\usepackage{xspace}
\usepackage{balance} 
\usepackage{csquotes}
\usepackage{makeidx}
\usepackage{blindtext}

\let\labelindent\relax
\usepackage{enumitem}

\usepackage{ragged2e}
\usepackage{soul} 
\usepackage{subfiles} 

\usepackage[protrusion=true,expansion=true]{microtype}
\usepackage[yyyymmdd]{datetime}

\date{\protect\formatdate{1}{1}{2001}}

\usepackage{url}
\makeatletter
\g@addto@macro{\UrlBreaks}{\UrlOrds}
\makeatother
\usepackage{color}
\usepackage[usenames,dvipsnames,table,xcdraw]{xcolor}
\usepackage{pgfplots} 
\pgfplotsset{compat=newest}

\usepackage{hyperref}
\hypersetup{
    colorlinks=true,
    linkcolor=black,
    citecolor=black,
    filecolor=cyan,
    urlcolor=black
}

\usepackage{marginnote}
\usepackage{soul} 

\usepackage{multicol}

\newcommand{\tocite}[1]{%
\textcolor{red}{[cite:\ifthenelse{\equal{#1}{}}{}{#1}?]}}

\newcommand{\ignore}[1]{}

\renewcommand{\baselinestretch}{1.0}



\makeatletter
\def\subsubsection{\@startsection{subsubsection}
                                 {3}
                                 {\z@}
                                 {0ex plus 0.1ex minus 0.1ex}
                                 {0ex}
                                 {\normalfont\normalsize\itshape}}
\makeatother

\usetikzlibrary{backgrounds}
\numberwithin{equation}{section} 

\tikzstyle{every node}=[font=\small]

\newcolumntype{L}[1]{>{\RaggedRight\hspace{0pt}}p{#1}}
\newcolumntype{R}[1]{>{\RaggedLeft\hspace{0pt}}p{#1}}

\renewcommand{\phi}{\varphi}

\newcommand{\furlp}[1]{\colorbox{blue!10}{\href{run:/home/fulong/academia/library/papers/#1.pdf}{D}}}
\newcommand{\furlb}[1]{\colorbox{blue!10}{\href{run:/home/fulong/academia/library/books/#1.pdf}{D}}}

\DeclareMathOperator{\arctantwo}{arctan2}

\renewcommand{\vec}[1]{\mathbf{#1}}

\newcommand{\phy}{\mathrm{p}}
\newcommand{\cmd}{\mathrm{c}}
\newcommand{\dpick}{\mathcal{D}_{\mathrm{pick}}}
\newcommand{\drand}{\mathcal{D}_{\mathrm{rand}}}

\ifCLASSINFOpdf
\else
\fi
\begin{document}
%
\title{Efficiently Calibrating Cable-Driven Surgical Robots with RGBD Fiducial Sensing and Recurrent Neural Networks}
%
%
%

\author{Minho Hwang$^{1}$, Brijen Thananjeyan$^1$,   Samuel Paradis$^1$, Daniel Seita$^{1}$, Jeffrey Ichnowski$^1$, \\ Danyal Fer$^2$,  Thomas Low$^3$, and  Ken Goldberg$^1$
\thanks{Manuscript received: March 15, 2020; Revised June 11, 2020; Accepted July 1, 2020.}
\thanks{This paper was recommended for publication by Editor Tamim Asfour upon evaluation of the Associate Editor and Reviewers' comments.} 
\thanks{$^{1}$Minho Hwang, Brijen Thananjeyan, Samuel Paradis, Daniel Seita, Jeffrey Ichnowski, and Ken Goldberg are with UC Berkeley.
        {\tt\footnotesize \{gkgkgk1215, bthananjeyan, samparadis, seita, jeffi, goldberg\}@berkeley.edu }}%
\thanks{$^{2} $Danyal Fer is with UCSF East Bay. {\tt\footnotesize danyal.fer@ucsf.edu }}%
\thanks{$^{3} $Thomas Low is with SRI International.  {\tt\footnotesize thomas.low@sri.com}}%
\thanks{Digital Object Identifier (DOI): see top of this page.}
}
%
%

\markboth{IEEE Robotics and Automation Letters. Preprint Version. Accepted July, 2020}
{Hwang \MakeLowercase{\textit{et al.}}: Efficiently Calibrating Cable-Driven Surgical Robots} 

%



\maketitle


\begin{abstract}
Automation of surgical subtasks using cable-driven robotic surgical assistants (RSAs) such as Intuitive Surgical's da Vinci Research Kit (dVRK) is challenging due to imprecision in control from cable-related effects such as cable stretching and hysteresis. We propose a novel approach to efficiently calibrate such robots by placing a 3D printed fiducial coordinate frames on the arm and end-effector that is tracked using RGBD sensing.
To measure the coupling and history-dependent effects between joints, we analyze data from sampled trajectories and consider 13 approaches to modeling.  These models include linear regression and LSTM recurrent neural networks, each with varying temporal window length to provide compensatory feedback. 
With the proposed method, data collection of 1800 samples takes 31 minutes and model training takes under 1 minute.
Results on a test set of reference trajectories suggest that the trained model can reduce the mean tracking error of physical robot from \SI{2.96}{\milli\meter} to \SI{0.65}{\milli\meter}.
Results on the execution of open-loop trajectories of the FLS peg transfer surgeon training task suggest that the best model increases success rate from \SI{39.4}{\percent} to \SI{96.7}{\percent}, 
producing performance comparable to that of an expert surgical resident. Supplementary materials, including code and 3D-printable models, are available at \url{https://sites.google.com/berkeley.edu/surgical-calibration} 
\end{abstract}

\begin{IEEEkeywords}
Calibration and Identification, Model Learning for Control, Medical Robots and Systems
\end{IEEEkeywords}

%
\IEEEpeerreviewmaketitle

%
%
%
%

\section{Introduction}
\IEEEPARstart{A}{ccurate} automated control of cable-driven surgical robots such as the da Vinci Research Kit (dVRK)~\cite{dvrk2014} is challenging due to cable-related effects such as hysteresis and cable tension~\cite{miyasaka2015,kalman_filter_2016}. These effects result in errors in the robot's odometry because the encoders that track joint configurations are frequently located near the motors but far from the joints. Automation of robot-assisted surgery can be very difficult due to the accuracy and precision required to perform surgical subtasks~\cite{minho_pegs_2020} and may require slow and tedious manual calibration~\cite{minho_pegs_2020,seita_icra_2018}. In current practical applications for robot surgery, human surgeons compensate for these inaccuracies.

\begin{figure}[t!]
\centering
\includegraphics[width=0.95\linewidth]{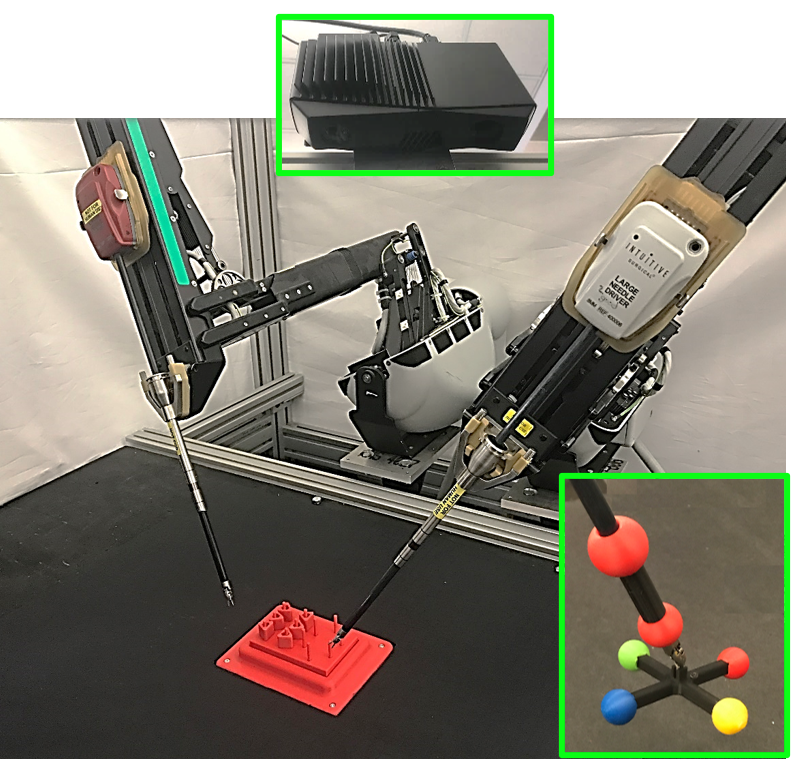}
\caption{
da Vinci Research Kit (dVRK) with 3D-printed spherical fiducials that facilitate tracking of true location of the six joints. A Zivid OnePlus RGBD camera is mounted 0.9 meters above the workspace (top).
} 
\label{fig:teaser}
\end{figure}

In this paper, we present a method to efficiently calibrate a dVRK that places 3D-printed fiducials on the end effector and arm (Fig.~\ref{fig:teaser}), uses RGBD images to estimate the robot's ground-truth joint configuration relative to its commanded joint configuration, trains a model, and implements a controller that compensates for history-dependent cabling effects. We also present an empirical analysis of these effects.

Prior work has studied learning to compensate joint estimation errors by considering current observations of robot state~\cite{mahler2014case, seita_icra_2018, hannaford_calibration_2019}.
To create a compensatory model for history-dependent effects, we incorporate a sequence of prior observations.
We investigate design choices such as temporal window size, linear modeling vs. deep learning, forward vs. inverse system identification, and usage of prior model outputs as current inputs. Empirically, the best model is a recurrent neural network on the forward dynamics with a temporal window of 4, which can reduce the mean-squared joint error by \SI{98.4}{\percent} on a held-out test set of random trajectory data. As the model fits data for a specific instrument to compensate for its specific cabling effects, the calibration procedure needs to be efficient. With the proposed method, data collection of 1800 samples takes 31 minutes and model training takes under 1 minute.

The accuracy of the ground-truth measurement system in this study is evaluated as 0.32 mm in sphere detection and less than 0.6 deg in joint angle estimation. The proposed calibration method could be adaptable to any depth-sensing or stereo cameras with easy-to-manufacture fiducials. The 3D-printed markers and the RGBD camera used in this study cost \$20 and \$1000 respectively.

We integrate the compensatory model into the robot's control loop and run automated trials of a variant of the Fundamentals of Laparoscopic Surgery (FLS) peg-transfer task using open-loop trajectories. The peg-transfer task
is a standard surgeon training task that requires high precision and dexterity to perform. Prior work on automating this task uses hand-tuned calibration procedures~\cite{minho_pegs_2020} or visual servoing~\cite{auto_peg_transfer_2015} to compensate for robot inaccuracies. Experiments suggest that the new calibration method can increase the per-block transfer success rate from \SI{39.4}{\percent} to \SI{96.7}{\percent} without visual servoing.

This work makes the following contributions: (1) a low-cost hardware fiducial design, used with RGBD to track ground-truth joint positions of the dVRK, (2) experiments suggesting that the wrist joints are the main source of state estimation errors, (3) correction of robot joint configuration estimation errors using an LSTM recurrent neural network that considers the history of observations and predictions, (4) physical robot experiments suggesting that the compensatory model can significantly increase the accuracy and precision of open-loop trajectories in a challenging peg-transfer task.

%

\section{Related Work}\label{sec:rw}
Surgeons routinely use surgical robots to perform operations through teleoperation; however, no procedure includes automated subtasks in clinical settings.
In research settings, several groups have demonstrated results in automating surgical subtasks, such as suturing~\cite{Schulman2013,sen2016automating,rosen_icra_suturing_2017, motion2vec}, cutting gauze~\cite{thananjeyan2017multilateral,rosen_icra_tissues_2019}, smoothing fabrics~\cite{seita_fabrics_2020}, identifying tumors via palpation~\cite{garg2016gpas}, performing debridement~\cite{Kehoe2014,murali2015learning}, tying knots~\cite{vandenBerg2010}, tissue manipulation~\cite{wang2018unified,alambeigi2018toward}, inserting and extracting surgical needles~\cite{automated_needle_pickup_2018,extraction_needles_2019,needle_insertion_deformation_2019}, and transferring blocks in the peg transfer task~\cite{auto_peg_transfer_2015,minho_pegs_2020}.

These surgical tasks often require absolute positional accuracies bounded within \SI{2}{\milli\meter}, which is difficult to obtain with cable-driven surgical-assistant robots ~\cite{dvrk2014}, ~\cite{raven2013}, ~\cite{hwang2017single}, especially flexible surgical robots~\cite{k-flex,hysteresis_effect}, as they are known to suffer from cable stretch, cable tension, and hysteresis. Research groups have taken various approaches to compensate for these inaccuracies, such as by using unscented Kalman filters to improve joint angle estimation~\cite{kalman_filter_2016} estimating cable stretch and friction~\cite{miyasaka2015}, or by learning offsets to correct for robot end-effector positions and orientations~\cite{pastor2013,mahler2014case,seita_icra_2018}.

Among the most relevant prior work, Peng~et~al.~\cite{hannaford_calibration_2019} recently reported a data-driven calibration method for the Raven II~\cite{raven2013}. They use three spheres and four RGB cameras to estimate the position of the end effector. They collect a labeled dataset of 49,407 poses including velocities and torques, and train a neural network to predict the 3D position to within \SI{1}{\milli\meter} accuracy on an offline test set. In contrast to their work, we consider the problem of estimating the joint configuration, which can be incorporated more directly in collision checking, and we also learn to predict the commanded input given a history and desired joint angle. Additionally, we use both RGB and depth sensing to track the sphere fiducials and consider historical motions in the estimation of joint angles, which enable compensation for hysteresis and backlash-like effects. Furthermore, we design practical controllers using these models and benchmark the result of applying the proposed calibration procedure on a challenging peg transfer task.


\section{Problem Definition}
\label{sec:problem_definition}

Let $\vec{q}_\phy$ be the specification of the degrees of freedom, or \emph{configuration}, of the da Vinci Research Kit.
Let $\mathcal{C}_\phy \subset \mathbb{R}^6$ be the set of all possible configurations, thus $\vec{q}_\phy \in \mathcal{C}_\phy$.
Let $\vec{q}_\cmd \in \mathcal{C}_\cmd$ be the robot's commanded configuration, where $\mathcal{C}_\cmd \subset  \mathbb{R}^6$ and is equivalent to the joint configurations measurable by the robot's encoders. 
Note that this differs from $\vec{q}_\phy$, because the encoders are located at the motors and away from the joints, which can result in a mismatch between the joint configuration measured by the encoders and the true joint configurations due to cabling effects.
We use subscripts to index specific joints in vectors, e.g., $\vec{q}^\top_\phy = \begin{bmatrix} q_{\phy, 1} & \ldots & q_{\phy,6} \end{bmatrix}$. See Fig.~\ref{fig:dh} for visualization of joints $q_1, \ldots, q_6$.  We suppress the ${}_\phy$ and ${}_\cmd$ subscripts when the distinction is not needed. Let $\tau_{t} = (\vec{q}^{(0)}_{\cmd}, \ldots, \vec{q}^{(t-1)}_{\cmd}) \in \mathcal{T}$ encode the prior trajectory information of the robot up to time $t$. In this work, we assume that the robot comes to a stop in between commands, due to the capture frequency of the Zivid depth camera used to estimate joint configurations (Section~\ref{ssec:pose_est}).
The goal of this paper is to compute functions:
\begin{align*}
f &: \mathcal{C}_\cmd \times \mathcal{T} \rightarrow \mathcal{C}_\phy \\
g &: \mathcal{C}_\phy \times \mathcal{T} \rightarrow \mathcal{C}_\cmd,
\end{align*}
where $f(\cdot)$ maps the current command at time $t$ and prior state information to the current physical state of the arm, and
$g(\cdot)$ maps the current physical state of the arm and its history to the command that was executed at time $t$.
The intent for $f(\cdot)$ is to determine the physical configuration of a robot given the commands that one sent to it.
The intent for $g(\cdot)$ is to derive commands to move the robot to a desired physical configuration.
At execution time, we would like to use the controller derived from $g(\cdot)$ to track a reference trajectory of target waypoints $(\vec{q}^{(t)}_{\rm d})_{t=0}^T$ where $\vec{q}^{(t)}_{\rm d} \in \mathcal{C}_\phy$. We also consider control generation via the forward model $f$ by approximately inverting it for a desired output waypoint.
Observe that these functions are history-dependent inverses of each other, i.e.,
$g(f(\vec{q}_{\textrm{c}}^{(t)}, \tau_t), \tau_t) = \vec{q}_{\textrm{c}}^{(t)}$ and
$f(g(\vec{q}_{\textrm{p}}^{(t)}, \tau_t), \tau_t) = \vec{q}_{\textrm{p}}^{(t)}$.
%

Complicating this objective is a significant sub-problem of computing $\vec{q}_\phy$ at any moment in time.
While $\vec{q}_\cmd$ can be readily determined by reading the encoders associated with each joint, $\vec{q}_\phy$ must be determined by other means. Additionally, because enumerating the set of possible trajectories is intractable, we estimate parametric approximations $f_\theta \approx f$ and $g_\phi \approx g$ from a finite sequence of samples $\mathcal{D} = \left((\vec{q}_{\cmd}^{(t)}, \vec{q}_{\phy}^{(t)})\right)_{t=0}^N$.

\section{Method}
\label{sec:method}
In this section, we describe methods to learn $f_\theta$ and $g_\phi$ from data and then use the learned models to more accurately control the robot. We start by sending a sequence of $\vec{q}_\cmd$ and tracking the physical trajectories of fiducial markers attached to the robot.  We then convert the marker's positions to $\vec{q}_\phy$ using kinematic equations.  After collecting several datasets of $\vec{q}_\cmd$ and its resulting $\vec{q}_\phy$, we then train 13 models and implement a controller.

\begin{figure}[t!]
\vspace*{3pt}
\centering
  \includegraphics[width=0.7\linewidth]{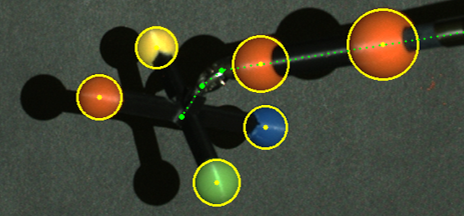}
  \caption{\textbf{Using sphere fiducials to estimate the pose of the end effector.} We place two spheres on the shaft to obtain the wrist position and four on the cross-shaped reference frame to find the orientation of the jaw. Yellow circles and dots indicate the detected spheres and their center locations. The green dotted lines show a skeleton of the estimated tool posture.}
 \vspace{3pt}
\label{fig:ball_fiducials}
\end{figure}

\subsection{Sphere Position Detection Algorithm}

To detect positions of spheres on the end effector, Peng~et~al.~\cite{hannaford_calibration_2019} uses 4 RGB cameras and OpenCV image processing functions, requiring 16 marker points to calibrate the cameras. In this work, to estimate $\vec{q}_\phy$, we use an RGBD camera to track fiducials of colored spheres attached to the end effector (Fig.~\ref{fig:ball_fiducials}). By masking depth and color ranges, we more robustly detect each of the spheres regardless of the image background. We attach 4 spheres on the end effector to ensure at least 3 are visible in case of occlusions. We mount 2 additional spheres on the tool shaft to decouple the first 3 joints from the last 3 and therefore to identify potential coupling effects between joints. This also reduces the effect of kinematic parameter inaccuracy on the joint estimation. From the derived kinematic equations in Section~\ref{ssec:pose_est}, note that $q_3$ is the only joint that is affected by the kinematic parameters $L_1$ and $L_\mathrm{tool}$ (Fig.~\ref{fig:dh}). We design and place the six spheres where they cannot overlap in the camera image within the working range of joints. We implement image segmentation using functions from OpenCV~\cite{opencv_library}.

With $n$ 3D points $(x_{i},y_{i},z_{i})_{i=0}^{n-1}$ corresponding to each segmented sphere obtained from the depth image, we formulate the following least squares estimate $A\vec{c} = \vec{b}$:
\begin{equation*} 
\underbrace{
\begin{bmatrix}
x_0 & y_0 & z_0 & 1 \\
\vdots & \vdots & \vdots & \vdots \\
x_{n-1} & y_{n-1} & z_{n-1} & 1 \\
\end{bmatrix}
}_{A \in \mathbb{R}^{n \times 4}}
\underbrace{
\begin{bmatrix}
c_0 \\
c_1 \\
c_2 \\
c_3
\end{bmatrix} 
}_{\vec{c} \in \mathbb{R}^{4\times 1}}
=
\underbrace{
\begin{bmatrix}
x_0^2 + y_0^2 + z_0^2  \\
\vdots \\
x_{n-1}^2 + y_{n-1}^2 + z_{n-1}^2
\end{bmatrix} 
}_{\vec{b} \in \mathbb{R}^{n\times 1}}
\end{equation*}
to obtain $\vec{c}$, and thus the center position $\vec{p}_b$ of each sphere $b$:
\begin{equation*}
\vec{p}_b= \begin{bmatrix} x_{b} & y_{b} & z_{b} \end{bmatrix}^\top = \frac{1}{2} \begin{bmatrix} c_0 & c_1 & c_2 \end{bmatrix}^\top,
\end{equation*}
with its radius as $r_{b}=\sqrt{c_3 + x_b^2 + y_b^2 + z_b^2}$.

\subsection{Configuration Estimation of Surgical Tool}\label{ssec:pose_est}

\begin{figure}[t!]
\vspace*{3pt}
\centering
\includegraphics[width=0.85\linewidth]{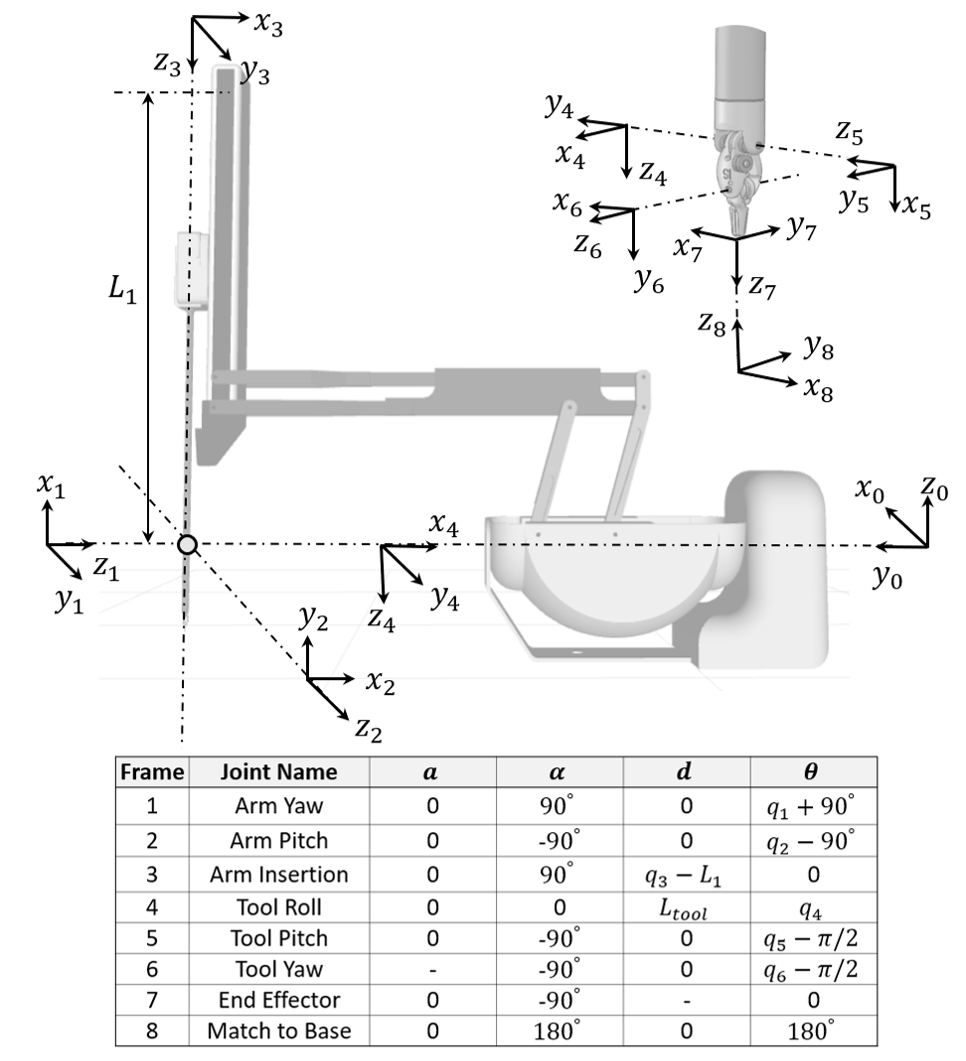}
\caption{\textbf{Coordinate frames using modified Denavit–Hartenberg convention.} For $i \in \{1, \ldots, 6\}$, the illustrated coordinate frame $(x_i, y_i, z_i)$ above corresponds to $q_i$ as used in the text. We use the kinematic equations described in Section~\ref{ssec:pose_est} to estimate the tool pose.} 
\vspace*{2pt}
\label{fig:dh}
\end{figure}

To estimate configuration of the surgical tool from the detected sphere positions, we define the kinematic parameters following the modified Denavit-Hatenberg convention in Fig.~\ref{fig:dh}. We can obtain the joint angles $q_{\phy,1}$, $q_{\phy,2}$, and $q_{\phy,3}$ in terms of the position of the wrist joint $^0_5\vec{p}$, which is measured by extending two positions of the spheres on the tool shaft:
%
\begin{equation*}
^0_5\vec{p} =
\begin{bmatrix}
^0_5x \\
^0_5y \\
^0_5z
\end{bmatrix} 
=
\begin{bmatrix}
\cos(q_{\phy,2})\cdot\sin(q_{\phy,1})\cdot(L_\mathrm{tool}-L_1+q_{\phy,3})\\
-\sin(q_{\phy,2})\cdot(L_\mathrm{tool}-L_1+q_{\phy,3})\\
-\cos(q_{\phy,1})\cdot\cos(q_{\phy,2})\cdot(L_\mathrm{tool}-L_1+q_{\phy,3})
\end{bmatrix},
\end{equation*}%
\vspace{-8pt}%
\begin{flalign*}
\text{which gives}\hspace{1em} &
\begin{bmatrix}
q_{\phy,1} \\
q_{\phy,2} \\
q_{\phy,3}
\end{bmatrix} 
=
\begin{bmatrix}
\arctantwo(^0_5x /-^0_5z)\\
\arctantwo(-^0_5y/\sqrt{^0_5x^2 + ^0_5z^2})\\
\sqrt{^0_5x^2 + ^0_5y^2 + ^0_5z^2} + L_1 - L_\mathrm{tool}
\end{bmatrix}, &
\end{flalign*}
where $L_\mathrm{tool}$ is the length of the surgical tool.

The last three joint angles, $q_4$, $q_5$, and $q_6$, can be computed from the rotation matrix $^{\hspace{0.3em}0}_\mathrm{fid}R$, which is obtained from the 4 fiducial spheres on the jaw as follows:
\begin{equation*}
\begin{bmatrix}
q_{\phy, 4} \\
q_{\phy, 5} \\
q_{\phy, 6}
\end{bmatrix} 
=
\begin{bmatrix}
\arctantwo(-r_{22}/r_{12})\\
\arctantwo(-r_{31}/r_{33})\\
\arctantwo\left( r_{32} / \sqrt{r_{31}^2 + r_{33}^2} \right)
\end{bmatrix},
\end{equation*}%
\vspace{-8pt}%
\begin{flalign*}
\text{where}\hspace{1em} &
\begin{bmatrix}
r_{11} & r_{12} & r_{13}\\
r_{21} & r_{22} & r_{23}\\
r_{31} & r_{32} & r_{33}\\
\end{bmatrix}
= \hspace{.2em} ^3_8R = (^0_3R)^{-1}\cdot^{\hspace{0.3em}0}_\mathrm{fid}R, &
\end{flalign*}
where $^i_jR$ is a rotation from frame $i$ to $j$. This gives us the full procedure for getting all six joints. 

\subsection{Data Collection}\label{ssec:data_collect}
To facilitate the data collection process, we first obtain the transformation from the RGBD camera to the robot. While the robot randomly moves throughout its workspace, we collect the wrist positions $^5\vec{p}_{org}$ using sphere detection and compare them with the positions reported by the robot. We use least squares estimation to match 684 sample points for the different base frames to obtain a $4x4$ transformation matrix. In the data collection process, we generate a training dataset containing pairs of desired and actual joint configurations of the robot. We randomly sample end-effector positions of the robot uniformly throughout a workspace of \SI{100}{\milli\meter} x \SI{80}{\milli\meter} x \SI{40}{\milli\meter}. 
We convert the positions to configurations for joints $q_1$, $q_2$, and $q_3$, and
randomly sample configurations for $q_4$, $q_5$, and $q_6$ within their respective joint limits. Then, we command the robot to replay the executed trajectory with the sphere fiducials attached.
This enables us to collect ground truth information for trajectories that are executed during a task. During the process, we collect the configuration $\vec{q}_\phy$ estimated from the fiducials and commanded joint angles $\vec{q}_\cmd$ to compile a dataset $\mathcal{D}$:
\begin{equation}
\mathcal{D} = \left((\vec{q}^{(t)}_{\cmd}, \vec{q}^{(t)}_{\phy})\right)_{t=1}^{N}.
\end{equation}

\begin{figure}[t!]
\vspace*{3pt}
\centering
\includegraphics[width=0.7\columnwidth]{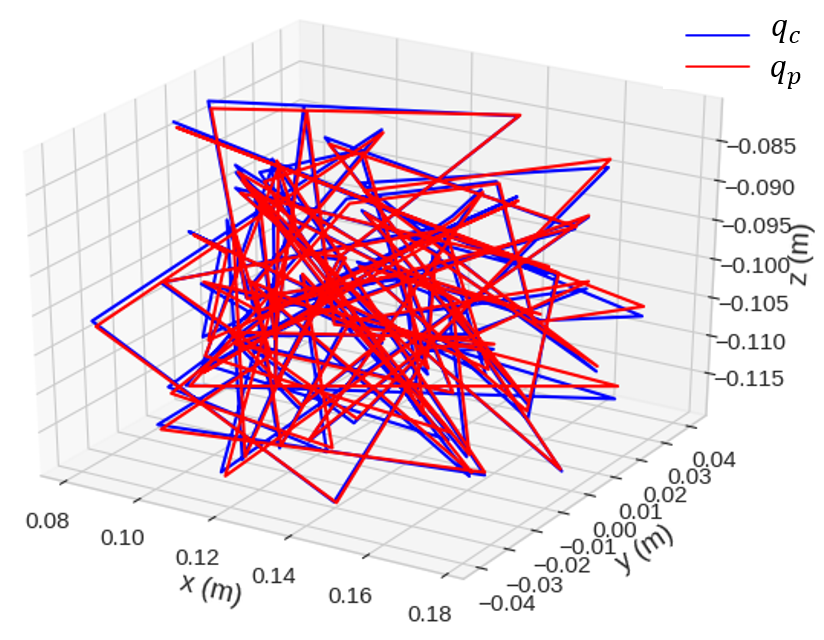}
\caption{\textbf{Random Trajectories:} We collect a dataset $\mathcal{D}_{\textrm{rand}}$ of randomly sampled configurations of the robot throughout its workspace for training. We collect both the commanded (desired) joint angles and the physical joint angles estimated from the fiducials.
} 
\vspace*{3pt}
\label{fig:random_data}
\end{figure}

\begin{figure}[t!]
\vspace*{3pt}
\centering
\includegraphics[width=0.7\columnwidth]{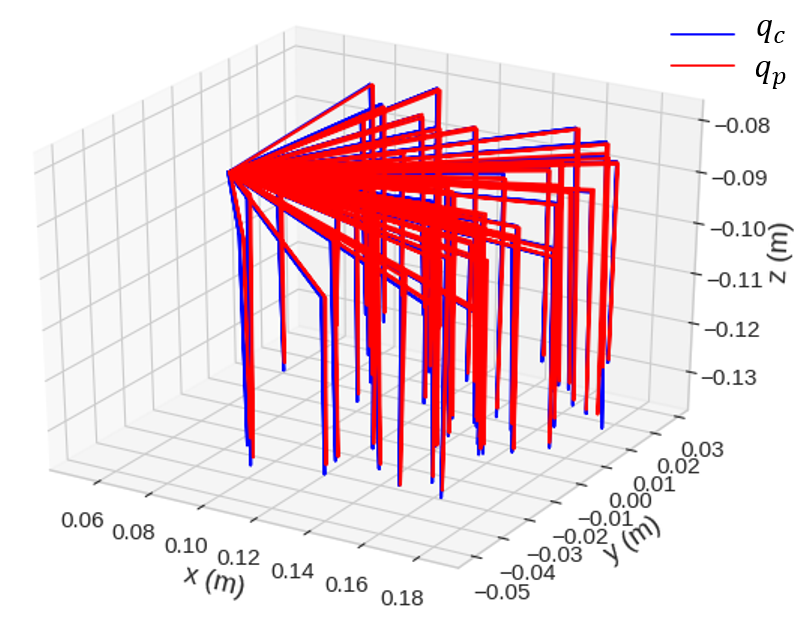}
\caption{\textbf{Pick and Place Trajectories:} We collect a dataset $\mathcal{D}_{\textrm{pick}}$ of pick and place motions in the workspace. We use these motions to train a controller specifically optimized for the peg transfer task (Section~\ref{subsec:peg_transfer}).}
\vspace*{3pt}
\label{fig:peg_data}
\end{figure}

We collect the following training datasets:
\begin{itemize}
    \item Random motions ($\mathcal{D}_{\textrm{rand}}$): This dataset consists of random sampled configurations of the robot (Fig.~\ref{fig:random_data}).  It takes 66 minutes to collect 4000 datapoints.
    \item Pick and place motions ($\mathcal{D}_{\textrm{pick}}$): This dataset consists of horizontal motions where the $z$ coordinate of the end effector is fixed and vertical motions where only the $z$ coordinate is varied (Fig.~\ref{fig:peg_data}). It takes 31 minutes to collect 1800 datapoints.
\end{itemize}
We collect test datasets in the same way that are \SI{10}{\percent} the size of the training datasets. Each dataset is collected over a single run on the robot.

\subsection{Error Identification}\label{ssec:error_id}
\begin{figure*}[t!]
\vspace*{3pt}
\centering
\includegraphics[width=0.83\linewidth]{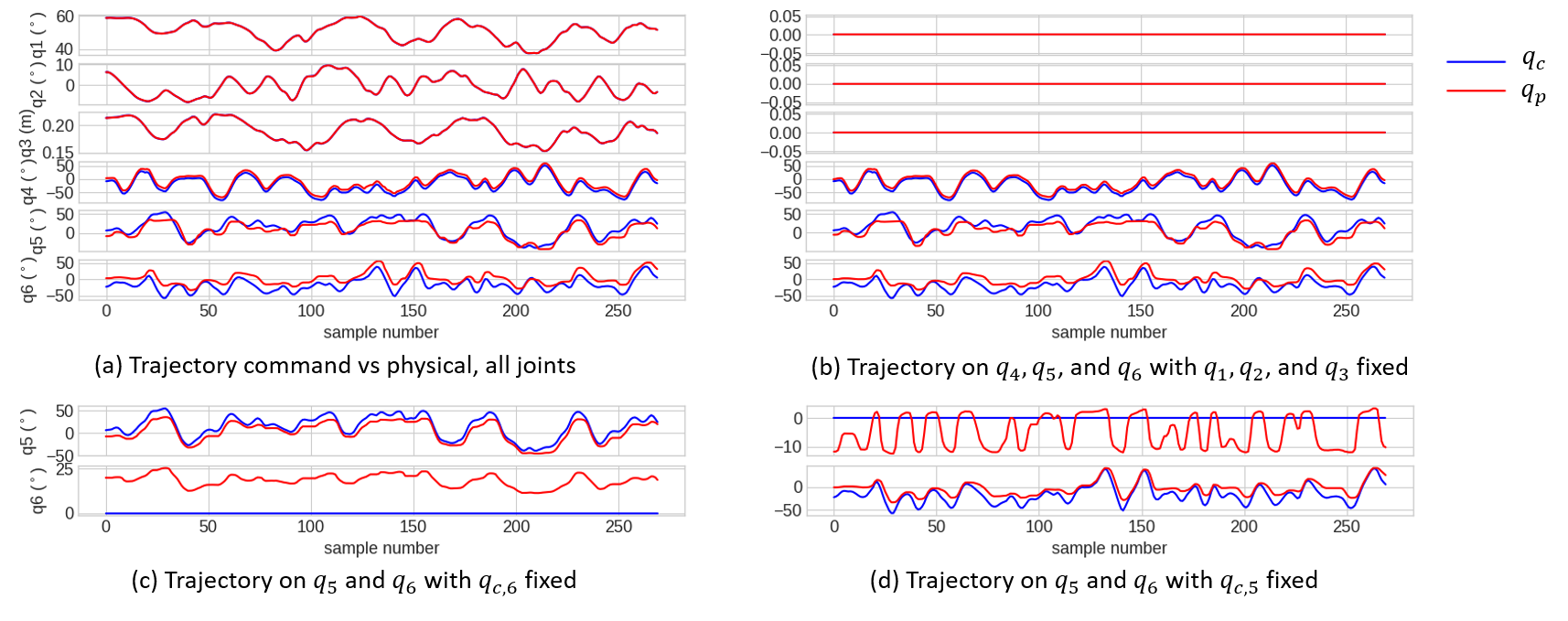}
\caption{\textbf{Top row:} We observe that the tool joint angles $q_{\phy, 4}$, $q_{\phy, 5}$, $q_{\phy, 6}$ are not affected by the movement of the external robot arm $q_{\phy, 1}$, $q_{\phy, 2}$, $q_{\phy, 3}$. The RMSE value are $[q_{\phy, 4}, q_{\phy, 5}, q_{\phy, 6}] = [10.7, 11.0, 20.6]$ and  $[10.0, 11.9, 19.9](\textrm{deg})$ before and after fixing the external arm. \textbf{Bottom row:} Two wrist joint angles, $q_{\phy, 5}$ and $q_{\phy, 6}$, are closely coupled to each other. Joint $q_6$ moved in correspondence to $q_{\phy, 5}$ despite its desired command was fixed as constant, and vice versa.
}

\vspace*{-2pt}
\label{fig:joint_trajectories}
\end{figure*}

We conduct a preliminary study to identify the characteristics of error and the coupling effect among joints. We sub-sample a portion of the dataset from Section~\ref{ssec:data_collect} with $N=270$. We then replay the trajectory but keep the first 3 joint angles fixed. Fig.~\ref{fig:joint_trajectories} presents the desired and measured trajectory of each joint angle in both cases. We notice that the three joints of the robot arm,  $q_{\phy, 1}$, $q_{\phy, 2}$, and $q_{\phy, 3}$, rarely contribute to the error compared to the last 3 joints, since the root mean square (RMS) errors are \SI{0.063} deg, \SI{0.049} deg, and \SI{0.255}{\milli\meter} respectively. The three joints of the surgical tool, $q_{\phy, 4}$, $q_{\phy, 5}$, and $q_{\phy, 6}$, are repeatable and not affected by the arm joints. In addition, results suggest that the last two joints are closely coupled, since $q_{\phy, 5}$ synchronously moved with $q_{\phy, 6}$ even though it was commanded to be stationary, and vice versa (Fig.~\ref{fig:joint_trajectories}). We hypothesize this occurs because these two joints have two additional cables that extend together along the shaft of the tool.

\subsection{State Estimation Without Attached Fiducials}
\label{sec:state-est-sans-fixtures}

To estimate 
$\vec{q}_\phy$ without the fiducials attached, we propose training a function approximator $f_\theta : \mathcal{C}_\cmd \times \mathcal{T} \rightarrow \mathcal{C}_\phy$, such that $f_\theta(\vec{q}_\cmd^{(t)}, \tau_t) = \hat{\vec{q}}_\phy^{(t)} \approx \vec{q}_\phy^{(t)}$.
We also 
train an inverse model $g_\phi(\vec{q}_\phy^{(t)}, \tau_t) = \hat{\vec{q}}_{\cmd}^{(t)}\approx \vec{q}_{\cmd}^{(t)}$, where $\theta$ and $\phi$ represent the parameters of the learned models.
In Section~\ref{subsec:positioning_performance}, we investigate training $f_\theta$ and $g_\phi$ using linear regression and deep neural networks of differing architectures and prior state inputs.

\begin{algorithm}[t]
  \caption{Control Optimization Algorithm}
  \label{alg:control-refine}
\begin{algorithmic}[1]
  \REQUIRE{Target position $\vec{q}^{(t)}_{\rm d}$, state estimator $f_\theta$, number of iterations $M$, learning rate $\alpha$}
  \STATE $\vec{q}^{(t)}_{\cmd} \leftarrow \vec{q}^{(t)}_{\rm d}$
  \FOR{$j \in \{1,\dots,M\}$}
    \STATE $\Delta_j \leftarrow \vec{q}^{(t)}_{\rm d} - f_\theta(\vec{q}^{(t)}_\cmd;\tau_t)$
    \STATE $\vec{q}^{(t)}_\cmd \leftarrow \vec{q}^{(t)}_\cmd + \alpha\Delta_j$
  \ENDFOR
  \RETURN $\vec{q}^{(t)}_\cmd$
\end{algorithmic}
\end{algorithm}

\subsection{Controller Design}
\label{subsec:controller_design}
Once we train function approximators, we would like to apply them to accurately control the robot while compensating for the robot's cabling effects.
Since the learned inverse model $g_\phi$ estimates commands, it can be used directly for this purpose.  However, as we show in Section~\ref{sec:experiment}, in some cases we can more accurately position the robot using the forward model as a basis for a controller.
This controller takes as input 
the target joint configuration $\vec{q}^{(t)}_{\rm d}$
and computes joint configuration command $\vec{q}^{(t)}_{\cmd}$ to get the robot to that configuration.
The algorithm for this controller (Alg.~\ref{alg:control-refine}) 
iteratively refines the command based on the error relative to the target position. The algorithm evaluates the forward dynamics $f_\theta$ for an input command to obtain an estimate of the next configuration $f_\theta(\vec{q}^{(t)}_\cmd;\tau_t)$. Then, the algorithm guides the commands to compensate for the error relative to the target position. This process is repeated for $M$ iterations.


\section{Experiments}
\label{sec:experiment}


We evaluate the performance of the proposed methods on a dVRK robot~\cite{dvrk2014}. The dVRK consists of two cable-driven, 7-DOF arms called patient-side manipulators (PSMs) that can be teleoperated by operating master tool manipulators (MTMs) or commanded programmatically. We use the Zivid OnePlus RGBD camera to track the sphere positions on the arm and end effector, which can provide 1920x1200 pixel images at 13 frames per second with depth resolution \SI{0.5}{\milli\meter}.

\subsection{Evaluation of Measurement System}
\label{subsec:measurement_performance}
To evaluate the accuracy of the measurement system, we fabricate 4 fiducial parts of different heights. We randomly configure them on precisely drilled surface plates at regular intervals. The measurement error is calculated by the difference between the sphere distances from the proposed detection method and the distances from the CAD model, which is considered as ground-truth. As a result of measuring 120 samples, we obtain a \SI[separate-uncertainty = true]{0.32 \pm 0.18}{\milli\meter} RMS error in single sphere detection. Within a range of the maximum error of \SI{0.67}{\milli\meter}, we add a uniform distribution of noise to the detection of six spheres and estimated joint angles of the robot. The results of the measurement resolution are summarized in Table~\ref{tab:experiment_results}. The accuracy of the measurement system is comparable to commercial 3D tracking systems including Aurora (\SIrange[range-phrase={--},range-units=single]{0.48}{0.70}{\milli\meter}) and Polaris (\SIrange[range-phrase={--},range-units=single]{0.25}{0.30}{\milli\meter}) from NDI Medical, and the detection using multiple RGB cameras (\SI{0.99}{\milli\meter})~\cite{hannaford_calibration_2019}.

\begin{table}[h]
\centering
\caption{\textbf{Accuracy of Measurement System}}
\resizebox{0.93\linewidth}{!}{
\centering
\renewcommand{\arraystretch}{1.5} 
\begin{tabular}{ P{0.7cm} | P{1.4cm} | P{0.7cm} P{0.7cm} P{0.5cm} P{0.7cm} P{0.7cm} P{0.7cm}}
\Xhline{2\arrayrulewidth}
&\textbf{Sphere} & \multicolumn{6}{c}{\textbf{Joint angle }}\\
\hline
&3D position (mm) & q1 (deg) & q2 (deg) & q3 (mm) & q4 (deg) & q5 (deg) & q6 (deg) \\
\hline
\hline
\textbf{RMS} & 0.32 & 0.022 & 0.022 & 0.42 & 0.34 & 0.35 & 0.59 \\
\textbf{SD} & 0.18 & 0.022 & 0.022 & 0.42 & 0.34 & 0.35 & 0.59 \\
\Xhline{2\arrayrulewidth}
\end{tabular}}
\vspace{0pt}
\label{tab:experiment_results}
\end{table}

\subsection{Modeling Approaches}
\label{subsec:positioning_performance}
In this section, we consider 13 learned models to estimate the ground-truth joint configuration and end-effector pose of the arm. We collect a large dataset of random motions and pick-and-place motions as described in Section~\ref{ssec:data_collect}. We also generate similar but smaller datasets as a validation set. All models are implemented in PyTorch~\cite{NEURIPS2019_9015} and trained using the MSE loss function and predict the three wrist joint angles $\hat{\vec{q}}^{(t)}_{\phy,4:6}$. The first three joints are not predicted or used as input, as we find they are very accurate and decoupled from the wrist joints as shown in Fig.~\ref{fig:joint_trajectories} and discussed in Section~\ref{ssec:error_id}.
We compute pose error by computing the forward kinematics of the arm using the new joint estimate.
We investigate the following modeling choices for the forward model $f_\theta$:
\subsubsection{Architecture}
We consider the following candidate architectures for $f_\theta$:
\begin{itemize}
    \item[] \textbf{Linear:} A linear regression from input to output using $\ell_1$ regularization.
    \item[] \textbf{FF:} A feed-forward neural network with two hidden layers, with each hidden layer having 256 units.
    \item[] \textbf{RNN:} An LSTM~\cite{hochreiter1997long} with 256 units, followed by a feed-forward neural network with two hidden layers of 256 units. Recurrent neural networks such as LSTMs maintain a hidden state, which we hypothesize may better capture history dependent errors.
\end{itemize}
\subsubsection{Input Format}
To capture the history-dependent nature of cabling effects, we track a history of horizon $H > 1$ and use it to form two possible input formats:

\begin{itemize}
    \item[] \textbf{Cmd:} ($\vec{x}_\cmd$) is the prior commanded joint positions.
    \item[] \textbf{Est:} ($\vec{x}_\mathrm{e}$) is the prior predictions and current desired joint position.
\end{itemize}

Thus, the input takes one of the following forms:
\begin{equation}
\vec{x}_\cmd = \begin{bmatrix}
\vec{q}_{\cmd,4:6}^{(t)} \\
\vec{q}_{\cmd,4:6}^{(t-1)} \\
\vdots \\
\vec{q}_{\cmd,4:6}^{(t-H)} \\
\end{bmatrix},
\qquad
\vec{x}_\mathrm{e} = \begin{bmatrix}
\vec{q}_{\cmd,4:6}^{(t)} \\
\hat{\vec{q}}_{\phy,4:6}^{(t-1)} \\
\vdots \\
\hat{\vec{q}}_{\phy,4:6}^{(t-H)} \\
\end{bmatrix}.
\end{equation}
We use $\vec{x}_t$ to denote a input in one of these forms. At training time, we use the ground-truth $\vec{q}_{\phy,4:6}^{(t)}$ in place of the estimated $\vec{\hat{q}}_{\phy,4:6}^{(t)}$ to avoid compounding modeling errors.
For the LSTM, each element of $\vec{x}_t$ is supplied individually in a sequence.


\subsubsection{Output Format}

\begin{itemize}
    \item[] \textbf{Abs:} $\hat{\vec{q}}^{(t)}_{\phy,4:6} = f_\theta(\vec{x}_t)$: the model directly predicts the absolute joint angles
    \item[] \textbf{Delta:} $\hat{\vec{q}}^{(t)}_{\phy,4:6} = f_\theta(\vec{x}_t) + \vec{q}^{(t)}_{\cmd,4:6}$: the model predicts the error of the robot's estimate.
\end{itemize}

For the inverse model $g_\phi$, we perform the same search where $\vec{q}^{(t)}_{\cmd}$ is replaced by $\vec{q}^{(t)}_{\phy}$ in the input and the model predicts $\vec{\hat{q}}^{(t)}_{\cmd}$. Because $\vec{q}_{\phy}^{(t)}$ is not predicted by this model, we only use the $\vec{x}_\cmd$ input format. We use the above codes sequentially to refer to specific models (e.g. RNN-Cmd-Delta).
For the linear model, we use a LASSO 
linear regression~\cite{Tibshirani94regressionshrinkage}.
and observe that regularization produces more stable results than vanilla least-squares.

\begin{figure}[t!]
\vspace*{3pt}
\centering
  \includegraphics[width=0.80\linewidth]{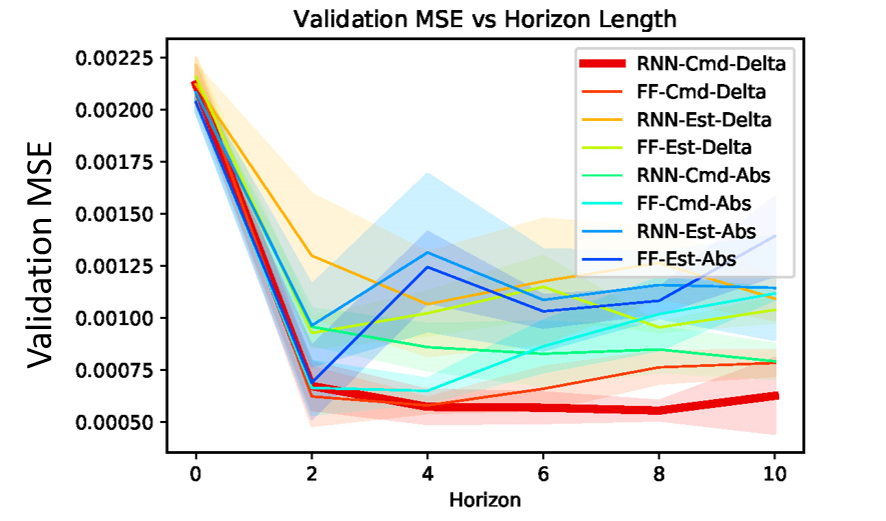}
  \caption{\textbf{Forward Model Horizon Ablation} In this experiment, we ablate the input horizon $H$ with respect to the models in Section~\ref{subsec:positioning_performance} on $\mathcal{D}_{\mathrm{pick}}$. We observe that the RNN model with the command as the input and predicted error as the output outperforms the other models, and roughly converges in performance around $H=4$. We observe that all models improve with a nonzero horizon, implying that temporal information is necessary to compensate for hysteresis, cable stretch, and backlash-like effects. We train each model $5$ times, reporting the mean and standard deviation.
  }
\vspace*{-3pt}
\label{fig:forward_horizon_ablation}
\end{figure}

\begin{figure}[t!]
\centering
  \includegraphics[width=0.80\linewidth]{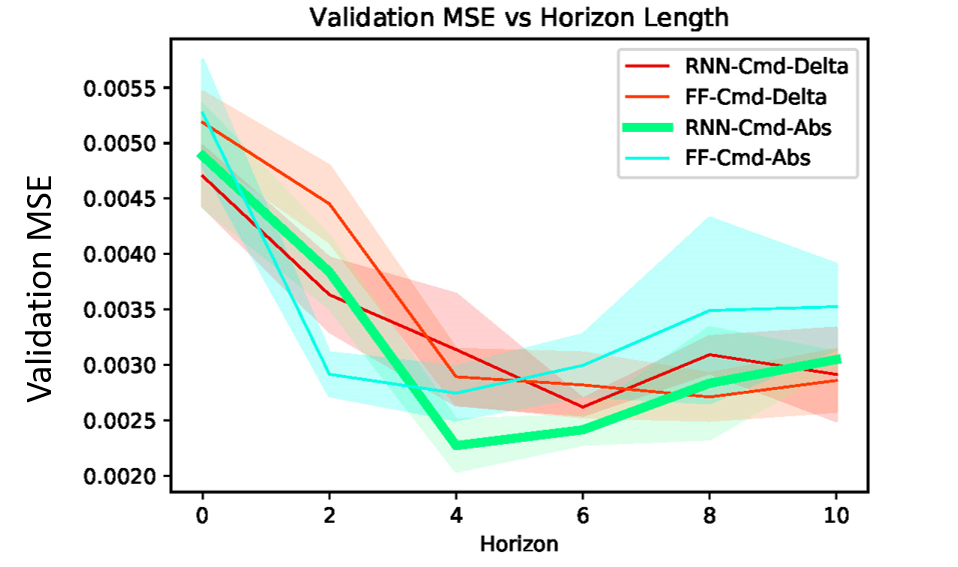}
  \caption{\textbf{Inverse Model Horizon Ablation} We ablate the input horizon $H$ with respect to the inverse versions of the models in Section~\ref{subsec:positioning_performance} on $\mathcal{D}_{\mathrm{pick}}$. We observe convergence in performance around $H=4$ using an RNN model with the command as the input and predicted joint angles as the output. We train each model $5$ times and report the mean and standard deviation.
  }
\vspace*{0pt}
\label{fig:inverse_horizon_ablation}
\end{figure}

\subsection{Offline Positioning Performance}
\label{subsec:offline_positioning}
We ablate all forward models in Section~\ref{subsec:positioning_performance} offline with respect to $H$ (Fig.~\ref{fig:forward_horizon_ablation}) and find that the validation mean-squared error (MSE) of most models start to converge around $H=2$, which suggests that some temporal information is necessary to compensate for cabling effects. We also ablate all inverse models considered in Section~\ref{subsec:positioning_performance} with respect to $H$ on $\dpick$ and find that the model 5 has the smallest validation MSE at $H=4$  (Fig.~\ref{fig:inverse_horizon_ablation}). Based on these results, we select model RNN-Cmd-Delta as the best forward model design and model RNN-Cmd-Abs as the best inverse model design. Hereafter, we refer to these specific models when we refer to the forward and inverse models. To increase robustness, we ensemble $10$ of these models when designing controllers.


\begin{figure}[t!]
\vspace*{3pt}
\centering
\includegraphics[width=\linewidth]{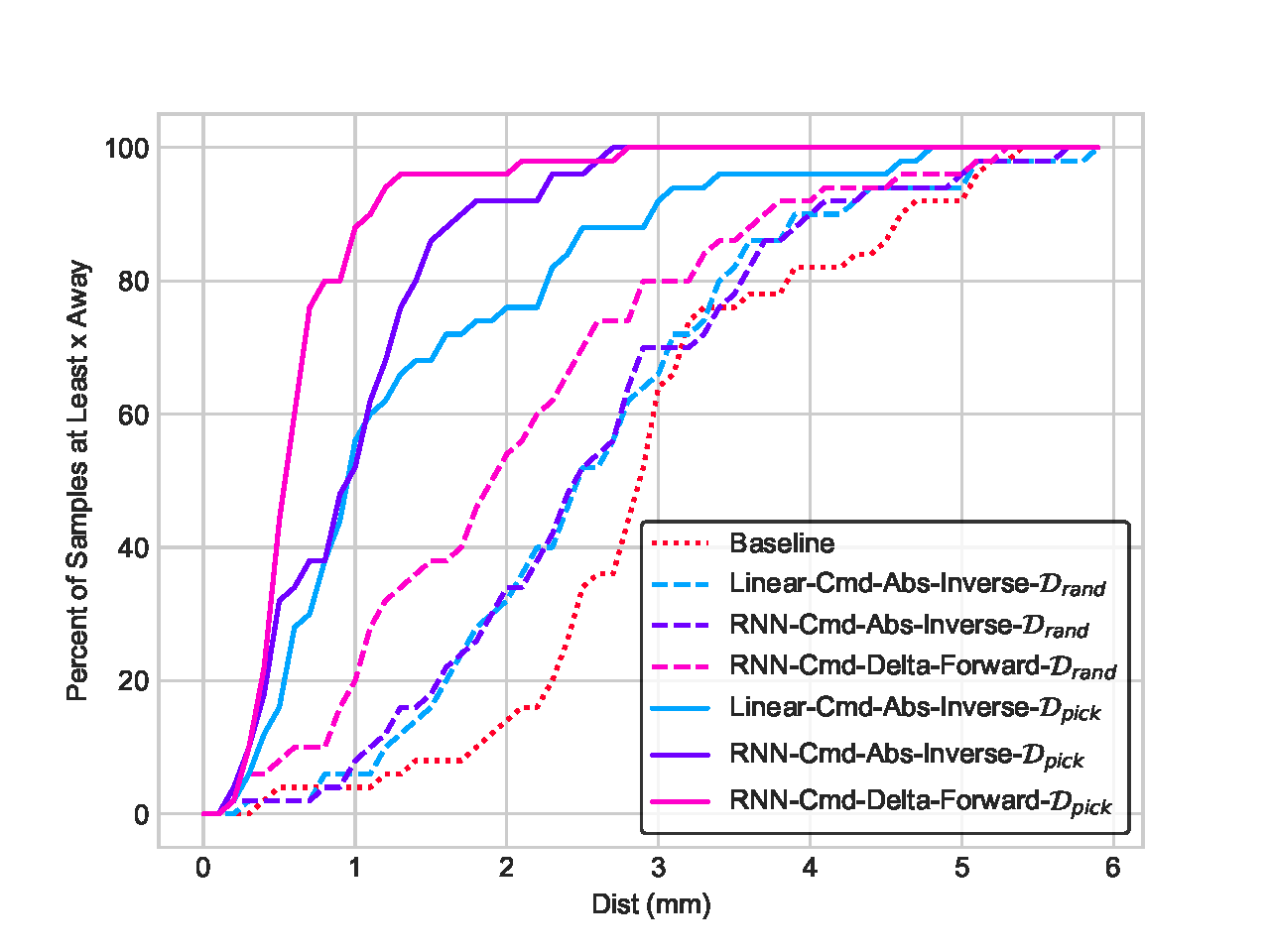}
\caption{\textbf{Cumulative Geometric Probability for Pick-and-Place Accuracy} Without the model, the end effector reaches within \SI{1}{\mm} of the desired physical location for \SI{4.0}{\percent} of pickups. With the model for correction, the robot ends within \SI{1}{\mm} desired location \SI{87.8}{\percent} of the time. For \SI{2}{\milli\meter}, this increases from \SI{14.1}{\percent} to \SI{95.9}{\percent}.
} 
\label{fig:traj_performance}
\end{figure}

\begin{table*}
\vspace*{5pt}
\centering
\caption{\textbf{Physical pick-and-place accuracy:} We compare $3$ different controller designs on $\drand$ and $\dpick$ and evaluate their accuracy on a sequence of previously unseen test pick-and-place motions on the dVRK. We observe that the forward model-based controller is the most accurate on average. Results suggest the mean tracking error is improved by training on trajectories that are similar to those executed during the task.}
\resizebox{0.7\linewidth}{!}{
\centering
\begin{tabular}{|l || c || c |  c || c | c || c | c |}
\hline
Cartesian Distances (mm) & Uncalibrated & \multicolumn{2}{c||}{Linear Inverse} & \multicolumn{2}{c||}{RNN Inverse} & \multicolumn{2}{c|}{RNN Forward}\\
\hline
Training Dataset & --- & $\drand$ & $\dpick$ & $\drand$ & $\dpick$ & $\drand$ & $\dpick$ \\
\hline\hline
Max & 6.20 & 5.90 & 4.74 & 5.68 & \textbf{2.61}& 5.29 & 2.79\\
\hline
Min & 0.51 & 0.21 & 0.20 & 0.13 & \textbf{0.11}& \textbf{0.11} & 0.18\\
\hline
Mean & 2.96 & 2.58 & 1.34 & 2.57 & 0.97 & 2.05 & \textbf{0.65}\\
\hline
Median & 2.89 & 2.48 & 0.95 & 2.47 & 0.92 & 1.90 & \textbf{0.55}\\
\hline
Standard deviation & 1.19 & 1.14 & 1.05 & 1.15 & 0.60 & 1.21 & \textbf{0.44}\\
\hline
\end{tabular}}
\label{tab:tracking}
\end{table*}

\begin{table*}
\centering
\caption{
\textbf{Peg-Transfer Task Experiments:} We evaluate the learned controllers on 10 trials of the handover-free FLS peg-transfer task. We observe that the baseline successfully transfers the block in \SI{39.4}{\percent} of its attempts, while the forward model trained on random trajectory data is able to successfully transfer the block in \SI{69.2}{\percent} of its attempts. The forward model trained on task-specific insertion data is able to complete \SI{96.7}{\percent} of attempted transfers. Because the task involves transferring 6 blocks to the other side and transferring them back, each block is transferred twice.
If a block's first transfer fails, it cannot be transferred again, so some runs may have fewer than 12 transfer attempts.
The task-specific forward model never makes an error in the first wave.
}\resizebox{0.9\linewidth}{!}{
\centering
\begin{tabular}{|l | c || c | c |  c | c || c | c | c |}
\hline
Model & Dataset & Mean Transfer Time (s) & Pick Failure & Stuck & Fall  & Success / Attempts & Success Rate (\%)\\
\hline\hline
Baseline & --- & \textbf{10.0} & 90 & 3 & 10 & 67/170 & 39.4\\
\hline
RNN Forward & $\drand$  & 12.9 & 58 & \textbf{1} & \textbf{6} & 146/211& 69.2 \\
\hline
RNN Forward & $\dpick$ & 13.6 & \textbf{0} & 2 & \textbf{6} & \textbf{232/240} & \textbf{96.7}\\
\hline
\end{tabular}}
\label{tab:dvrk-results}
\end{table*}

\subsection{Trajectory-Tracking Task}
In this experiment, we evaluate the controller described in Section~\ref{subsec:controller_design} on a trajectory-tracking task, where the objective is to guide the robot along a desired target trajectory $(\vec{q}^{(t)}_{\rm d})_{t=0}^T$, by commanding it to the current timestep's waypoint. We evaluate the performance of various controllers on a sequence of unseen pick and place trajectories. The controllers are trained on either $\drand$ or $\dpick$ to measure the effect of distributional mismatch between the training data and target trajectories. For the forward model, we use $\alpha=0.5$ and $M = 3$. Because we only calibrate the wrist joints, we evaluate using ground-truth information for $\vec{q}_{\phy,1:3}$. We find that the compensatory model can reduce the mean tracking error of the physical robot from \SI{2.96}{\mm} to \SI{1.90}{\mm} training on $\drand$, and to \SI{0.65}{\milli\meter} training on $\dpick$ (Table~\ref{tab:tracking}).

\begin{figure}[t!]
\centering
\includegraphics[width=1.0\linewidth]{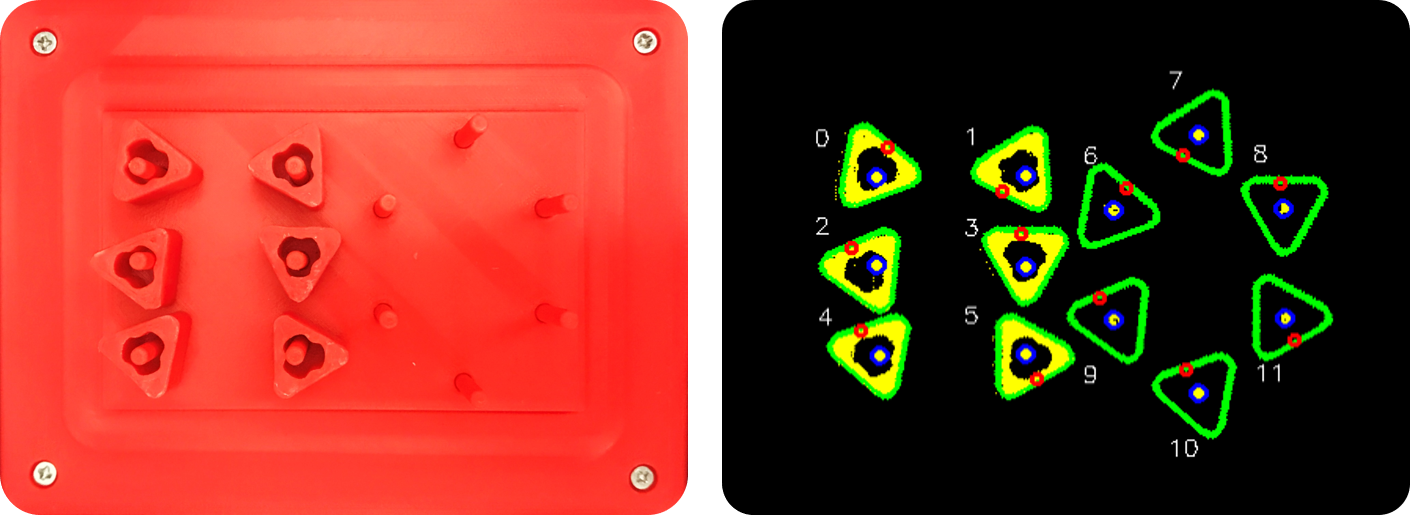}
\caption{Left: Blocks, pegs, and pegboard are all red to simulate a surgical setting. Blocks are detected using depth masking. Right: Detected blocks are shown in yellow and bordered green, while grasp points are shown in red. The bordered green without block indicates the estimated placing pose.
} 
\label{fig:peg_transfer}
\end{figure}

\subsection{FLS Peg Transfer Task}
\label{subsec:peg_transfer}
We evaluate the proposed approach on a variant of the 
FLS
peg-transfer task, in which the robot transfers 6 blocks from pegs on the left side of a pegboard to pegs on the right side, and then transfers them back (Fig.~\ref{fig:peg_transfer}). 
We do not consider arm-to-arm handover from the original task.
We use the monochrome setup from Hwang~et~al.~\cite{minho_pegs_2020} with 2 changes: 
(1) we 3D print the board red instead of painting it red to avoid issues with sticking pegs 
and (2) 
we install springs at the bottom of the board to avoid potential hardware damage to the dVRK. 
Incidental contact with the pegboard causes a change in pose of the pegs, so after each pick-and-place attempt, we capture a new RGBD image and regenerate the trajectory.
We evaluate the uncalibrated robot, forward-model controller, inverse-model controller, and linear-inverse-model controller on this task. We find that without calibration, the robot succeeds on \SI{39.4}{\percent} of individual transfers. With calibration on random trajectory data, the robot succeeds on \SI{69.2}{\percent}. With calibration on pick-and-place data, the robot succeeds on \SI{96.7}{\percent}.
It is promising that using calibration results in a success rate that is comparable to that of an expert human surgical resident (\SI{95.8}{\percent})~\cite{minho_pegs_2020} on the FLS peg-transfer task with a different setup.
This also suggests 
matching the distribution of trajectories used for training with the distribution encountered at test time can yield better compensation (Table~\ref{tab:dvrk-results}).



\subsection{Linear Regression Analysis}
\label{sec:disc}
\begin{figure}[t]
\centering
\includegraphics[width=1.0\linewidth]{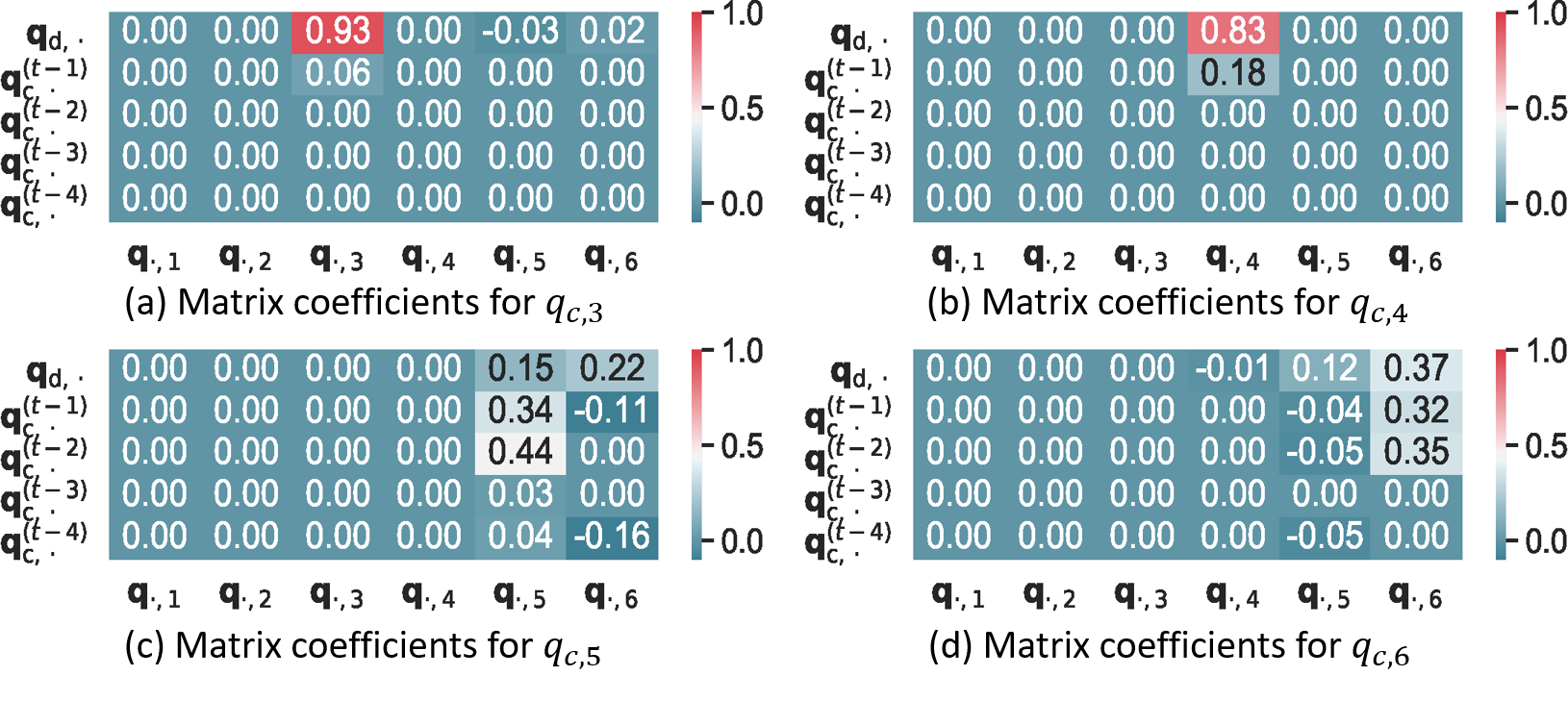}
\caption{\textbf{View of the $A$ matrix from linear regression.}  Each table shows a single row of $A$ reorganized to visualize the command and history effects on output command, with the target command in the first row, and the history of commands in subsequent rows.
Rows of $A$ for joints $q_1$ and $q_2$ are not included as they contain a single $1.0$ in mapping $q_{d,\cdot}$ to $q_{c,\cdot}$, similar to the $0.93$ for $q_3$ in (a).
Joints $q_4$ (b), $q_5$ (c), $q_6$ (d) experience hysteresis which is observable by the effect previous commands on the output.  Joints $q_5$ (c) and $q_6$ (d) are coupled, shown by their effects on each other.}
\label{fig:amatrix}
\end{figure}

To gain insight into the cabling effects on joint coupling and hysteresis, we inspect the linear regression model.
This model takes the form $g_\mathrm{lin}(\vec{x}_t) = A \vec{x}_t + \vec{b}$.  Fig.~\ref{fig:amatrix} shows the values in $A$ from a LASSO regression on a training dataset.  From the matrix, we observe that joints 1, 2, and 3 depend only on the target angle for that joint and not on another command or history.
Joint 4 exhibits some hysteresis as $q_{\cmd,4}$ is function of both the $q_{d,4}$ and the command history.
Joints $q_{5}$ and $q_{6}$ exhibit both hysteresis and joint couplings as their $\vec{q}_\cmd$ is tied to the $\vec{q}_d$ and the command history of both joints.  The linear regression results match our observations, and suggest why history-dependent models do better than baseline.



\section{Conclusion and Future Work}

We presented an efficient method for calibrating to the hysteresis and coupling effects of cable-driven surgical robots.  The method tracks fiducial markers with an RGBD camera to build a dataset from which we learn a model.  With a controller based on this model, we perform the FLS peg-transfer task with a success rate that suggests that this method can improve precision over baseline.

In future work, we will apply the proposed calibration method to improve accuracy for other surgical tasks such as cutting, debridement, and peg transfer with hand-overs. One limitation of this work is that it requires a dataset of trajectories for a specific arm and instrument. To overcome this limitation, we will investigate the application of meta learning~\cite{maml} to rapidly adapt to a new instrument or arm using a limited dataset by leveraging a large meta-training dataset generated from a set of arms and instruments.  While we did not notice issues with drift over time so far, we will also study how the robot's dynamics change over longer periods of time.
We will also investigate using the method to calibrate two PSM arms for multilateral surgical manipulation tasks.

\section*{Acknowledgements}
\small{
This research was performed at the AUTOLAB at UC Berkeley in affiliation with the Berkeley AI Research (BAIR) Lab, Berkeley Deep Drive (BDD), the Real-Time Intelligent Secure Execution (RISE) Lab, the CITRIS ``People and Robots'' (CPAR) Initiative, and with UC Berkeley's Center for Automation and Learning for Medical Robotics (Cal-MR). This work is supported in part by the Technology \& Advanced Telemedicine Research Center (TATRC) project W81XWH-19-C-0096 under a
medical Telerobotic Operative Network (TRON) project led by SRI
International. The authors were supported in part by donations from Siemens, Google, Toyota Research Institute, Honda, Intel, and Intuitive Surgical. The da Vinci Research Kit was supported by the National Science Foundation, via the National Robotics Initiative (NRI), as part of the collaborative research project ``Software Framework for Research in Semi-Autonomous Teleoperation'' between The Johns Hopkins University (IIS 1637789), Worcester Polytechnic Institute (IIS 1637759), and the University of Washington (IIS 1637444).  Daniel Seita is supported by a National Physical Science Consortium Fellowship. 
}

\ifCLASSOPTIONcaptionsoff
  \newpage
\fi



%



\bibliographystyle{IEEEtran}
\bibliography{IEEEabrv, reference}

%








\end{document}